\documentclass[11pt]{article}
\usepackage{coling2020}
\usepackage{times}
\usepackage{url}
\usepackage{latexsym}
\usepackage{subfigure}
\usepackage{enumitem}
\usepackage{microtype}
\usepackage{multirow}
\usepackage{graphicx}
\usepackage{booktabs}
\usepackage{amsmath}
\usepackage{amsfonts}
\usepackage{float}
\setenumerate[1]{itemsep=0pt,partopsep=0pt,parsep=\parskip,topsep=0pt}
\setitemize[1]{itemsep=0pt,partopsep=0pt,parsep=\parskip,topsep=0pt}
\setdescription{itemsep=0pt,partopsep=0pt,parsep=\parskip,topsep=0pt}

\colingfinalcopy 

\title{Subjective Bias in Abstractive Summarization}

\author{\textbf{Lei Li}$^{1}$,
\textbf{Wei Liu}$^{1}$,
\textbf{Marina Litvak}$^{2}$,
\textbf{Natalia Vanetik}$^{2}$,
\textbf{Jiacheng Pei}$^{1}$,
\textbf{Yinan Liu} $^{1}$,
\textbf{Siya Qi} $^{1}$
\\
$^{1}$Beijing University of Posts and Telecommunications \\
$^{2}$Shamoon College of Engineering \\
\tt
\normalsize{\{leili, thinkwee, lyinan, qsy\}@bupt.edu.cn katching.pei@gmail.com} \\ 
\tt
\normalsize{litvak.marina@gmail.com natalyav@sce.ac.il}}
\date{}

\begin{document}
\maketitle
\begin{abstract}
  Due to the subjectivity of the summarization, it is a good practice to have more than one gold summary for each training document. However, many modern large-scale abstractive summarization datasets have only one-to-one samples written by different human with different styles. The impact of this phenomenon is understudied. We formulate the differences among possible multiple expressions summarizing the same content as subjective bias and examine the role of this bias in the context of abstractive summarization. In this paper a lightweight and effective method to extract the feature embeddings of subjective styles is proposed. Results of summarization models trained on style-clustered datasets show that there are certain types of styles that lead to better convergence, abstraction and generalization. The reproducible code and generated summaries are available online.
\end{abstract}

\section{Introduction}
\label{intro}
Given a verbose input article, abstractive summarization aims at generating summaries covering its key facts. The base architecture to solve this problem is the attention-based encoder-decoder~\cite{rush2015neural} which greatly improved the result of neural translation~\cite{bahdanau2014neural}. Former studies proposed various models to better understand document~\cite{nallapati2016abstractive}, handle the out-of-vocabulary(OOV) problem~\cite{see2017get}, reduce the repetition\cite{chen2016distraction-based,li2019in} or divided the summarization problem into two steps(select and rewrite)~\cite{moroshko2019an,chen2018fast}. Recent researches also introduced Pretrained Language Model(PLM)~\cite{liu2019text,lewis2019bart:,raffel2019exploring} to this task. In our opinion, the abstractive summarization task is not only about identifying the key content of articles but also about Natural Language Generation(NLG) for summaries. Even the encoder caught and encoded the proper part of an article, the generated summaries may still be various depending on different human-written gold summaries. The summaries written by human are subjective and are therefore susceptible to recall bias. Generation bias in the dataset brought by human annotators should matter in the abstractive summarization task.

To study this problem, we hypothesize that there exists some subjective style bias among different samples. We define the writing style of human annotators formulating summaries after they have read and captured the main idea of articles as "Subjective Style". Our results on the most used CNN-Daily Mail(CNN-DM) dataset~\cite{hermann2015teaching} show that different style has a different impact on the model adaption, convergence speed, readability, and abstraction of generated summaries. In particular, this paper makes several contributions as follows:
\begin{itemize}
\setlength{\itemsep}{0pt}
\setlength{\parsep}{0pt}
\setlength{\parskip}{0pt}
\item The hypothesis about subjective bias among different samples in datasets is proposed, studied, and verified in detail for the first time with regard to abstractive summarization tasks on the CNN-DM dataset and its influence on the quality of the NLG process for summary generation is also evaluated. 
\item There are few related works about embedding writing style in the sequence-to-sequence(seq2seq) task in which subjective style can be seen as a special case. We propose to use the graph structure to represent the syntactic information in texts and put forward a self-supervised task to extract and embed the subjective style utilizing Graph Convolutional Network (GCN). 
\item Experimental results on style-clustered datasets confirm our assumption. Combining all styles in a dataset may not be the best practice for training abstractive summarization models.
\end{itemize}


\section{Problem Definition}
\definecolor{spancommon}{RGB}{31,123,187}
\begin{table*}[htbp]
    \footnotesize
    \centering
    \footnotesize
        \begin{tabular}{p{15cm}}
        \hline
          \textbf{Gold Summary 1}: \textcolor{spancommon}{liana barrientos married ten men in eleven years - even marrying six of them in one year alone. all of her marriages took place in new york state. her first marriage took place in 1999, followed by two in 2001, six in 2002, and her tenth marriage in 2010 .} liana barrientos allegedly described her 2010 nuptials as ` her first and only marriage ' she is reportedly divorced from four of her ten husbands. \\ \hline
          \textbf{Gold Summary 2}: \textcolor{spancommon}{liana barrientos married 10 men in 11 years - with six in one year alone. alleged scam occurred between 1999 and 2010 .} her eighth husband was deported back to pakistan for making threats against the us in 2006 after a terrorism investigation. the bronx woman plead not guilty to two fraud charges friday.  \\ \hline
          \textbf{Gold Summary 3}: liana barrientos, 39, re-arrested after court appearance for alleged fare beating. \textcolor{spancommon}{she has married 10 times as part of an immigration scam}, prosecutors say. liana barrientos pleaded not guilty friday to misdemeanor charges.   \\ \hline
        \end{tabular}
    \caption{An example of Subjective Bias in CNN-DM dataset. Here present only the summary part of three article-summary samples in training set. The ideal situation should be using different article-summary pairs with consistent style to build the dataset. But more often the dataset consists of different samples with different styles(even multiple expressions for the same fact just like this case)}
    \label{table:sample}
\end{table*}
Table \ref{table:sample} illustrates an example of three different gold summaries in the test dataset of CNN-DM describing the same content. These summaries use different syntactic structures and compression ratios to describe the fact "bigamy" which indicates that different styles do exist in one dataset. This paper will study the performance of deep abstractive summarization models trained on datasets with different styles or combined ones.

Take Table \ref{table:sample} for example, there are three summaries in the table and three corresponding articles which do not present here. Highlighted blue is the some sentences in summaries describing the content "bigamy" which comes from some parts of the article. We define the sentence in the article which describes the bigamy as \textbf{Oracle} and the corresponding summary sentence(in light blue) as \textbf{Summary} respectively. The \textbf{Subjective Style} should be a pattern of syntactic transformation from Oracle to Summary like deleting some adverbs and adjectives or compressing some subordinate clauses. Inconsistent Subjective Styles during the training phrase cause \textbf{Subjective Bias}. We propose to train a summarization model on a particular style and see different styles' impact on the results. So we extract \textbf{Subjective Style Embedding} of each article-summary sample then cluster samples with similar style to formulate datasets with one particular style. Hence our experiments are conducted in three steps:
\begin{itemize}
\setlength{\itemsep}{0pt}
\setlength{\parsep}{0pt}
\setlength{\parskip}{0pt}
\item Obtain the Subjective Style Embedding of each article-summary sample. 
\item Cluster the dataset based on the Subjective Style Embedding. 
\item Train different abstractive summarization models on clustered datasets and analyze the results.
\end{itemize}

\section{Method}
As mentioned before Subjective Style should be a pattern of syntactic transformation from Oracle to Summary, so we embed Subjective Style in the following steps:use SynGraph to represent syntactic information of sentence; set a task named LTRS to learn the syntactic embeddings of Oracle and Summary sentence considering their transformation; concatenate the Oracle syntactic embedding and Summary syntactic embedding as Subjective Style Embedding. After obtaining the embeddings we then cluster and divide datasets and train summarization models.
\subsection{SynGraph}
\begin{figure}[htbp]
  \centering
  \begin{minipage}[t]{0.5\linewidth}
  \centering
  \includegraphics[scale=0.18]{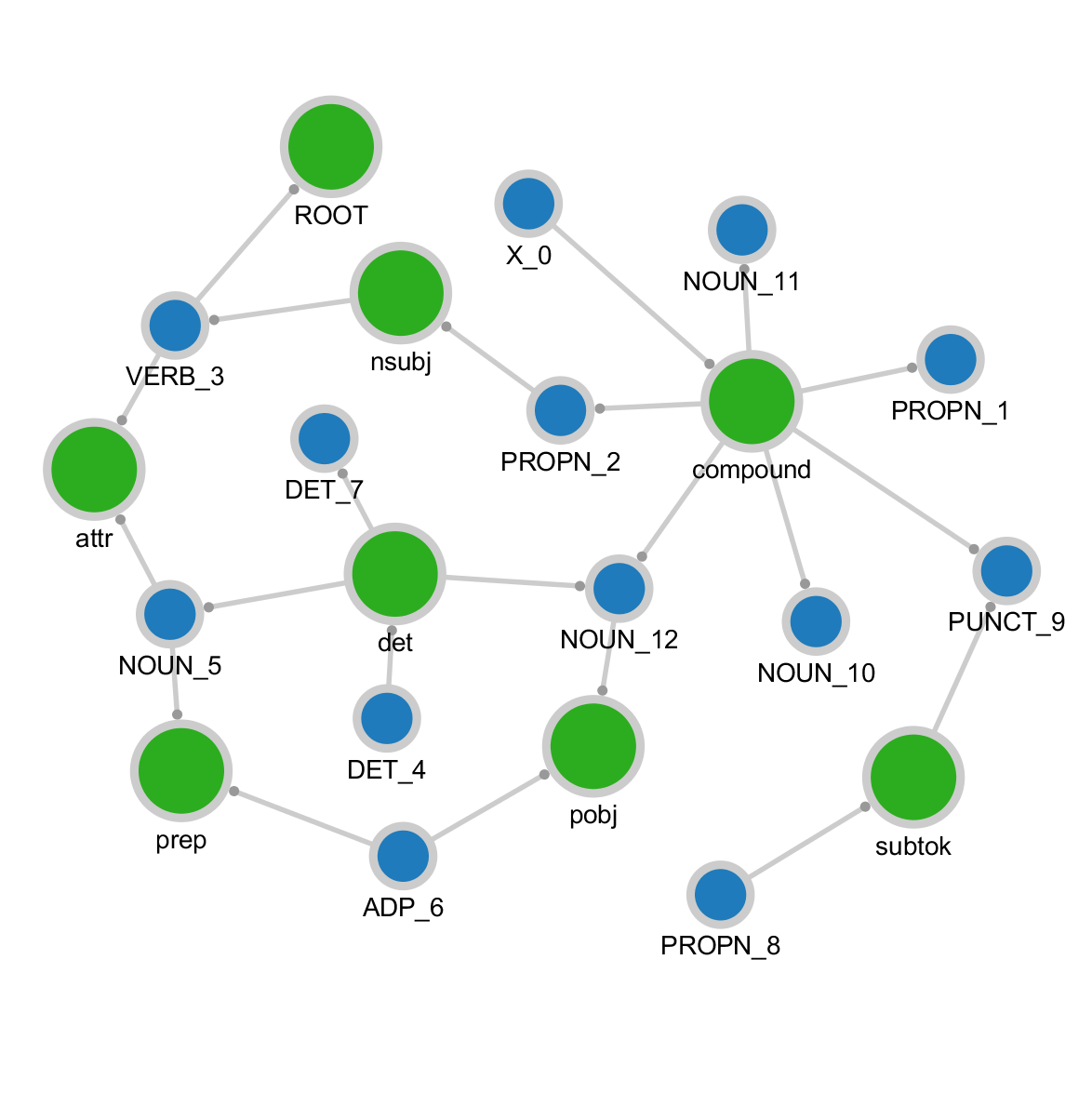}
  \caption{SynGraph}
  \label{fig:syn_graph}
  \end{minipage}%
  \begin{minipage}[t]{0.5\linewidth}
  \centering
  \includegraphics[scale=0.15]{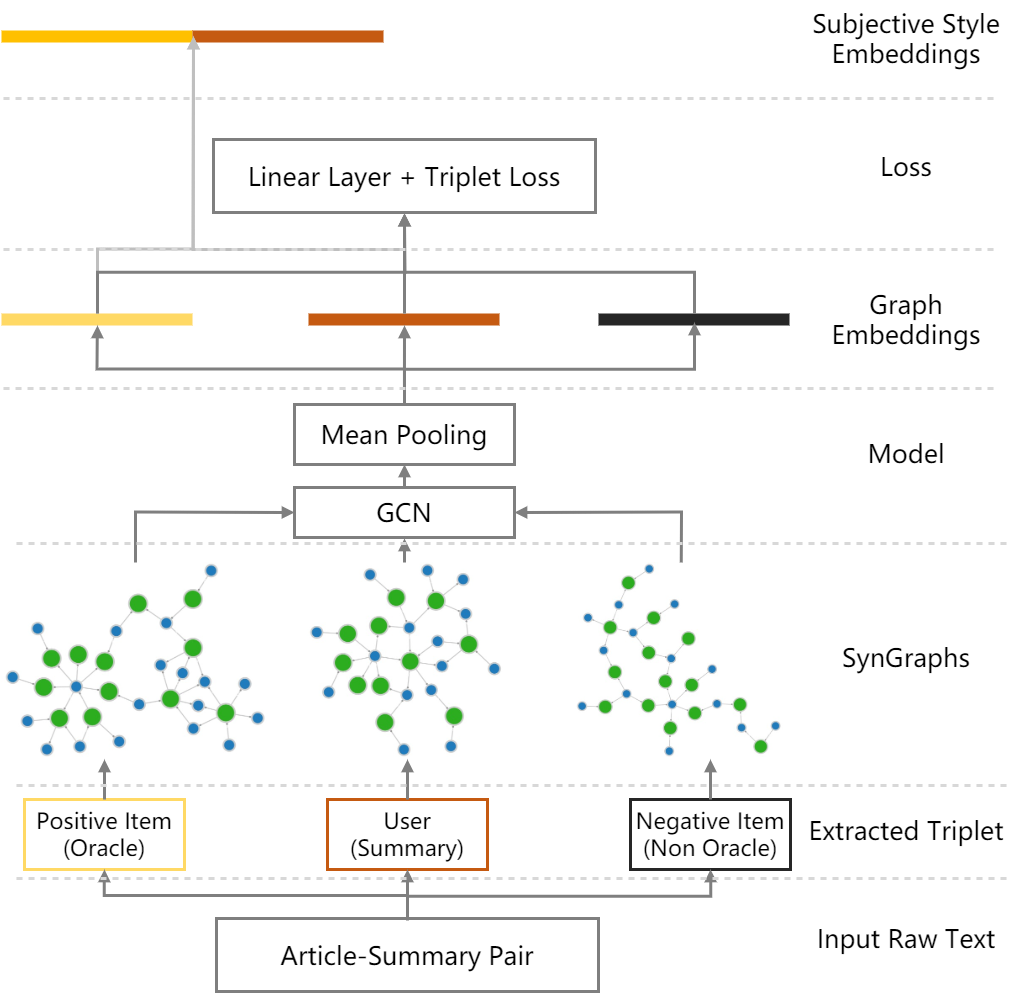}
  \caption{Overview of LTRS task}
  \label{fig:SubjGCN}
  \end{minipage}%
\end{figure}

First, to better embed the syntactic information of the sentence we construct a structure named SynGraph which is the graph form of dependency parsing results. Each sentence is represented as a graph. Words are treated as nodes and syntactic dependency relationships between words are regarded as edges. To focus on the grammatical structure instead of semantics, only part of speech(POS) is used to represent words. But it brings a problem: the edges are heterogeneous and hard to process. Considering that the node vocabulary is relatively small compared to the number of edge types, there is no need to construct a heterogeneous model. So the heterogeneous graph is transformed into a homogeneous one. Inspired by "Levi"~\cite{beck2018graph-to-sequence} operation, each dependency edge is changed into a node and inserted it between nodes sharing this dependency relationship. Then, all dependency nodes with the same label are merged. Figure \ref{fig:syn_graph} provides an example of SynGraph. We build both directed and undirected (bidirectional) versions for each graph were built.
\subsection{LTRS Task}
SynGraph makes extracting the syntactic embedding a task for embedding a graph. Since there is no supervised signal on the graph, we develop a self-supervised task to create supervised training samples from the original summarization dataset. The idea is just like Learning To Rank(LTR) task in the recommendation. For each user(Summary), model will try to rank all candidate items(all sentences in the article) for recommendation and the Oracle which shares the same content/fact with Summary is the best candidate item. Then the user embedding(Summary embedding) and item embedding(Oracle embedding and Non-Oracle embedding) can be obtained. As such we can train graph embeddings considering both the difference among samples and the connection between Summary and Oracle. Models are trained in the pair-wise way which makes the score between Summary-Oracle pair larger than Summary-Non Oracle pair. It is noteworthy that in our scenario both "user" and "item" are texts and both need an encoder for extracting SynGraph features. Thus all parts share a Graph Convolutional Network(GCN) encoder and use a linear layer to get the final score instead of simply calculating the embedding cosine similarity. This can prevent the mode collapse that Summary embeddings are almost the same as Oracle embeddings. We denote this task by Learning To Rank for Summarization(LTRS).

For each original article-summary sample, the first summary sentence is chosen as Summary(\textbf{user}) and approximate Oracle sentence(\textbf{positive item sample}) by finding the most Jaccard-similar sentence in the article to the Summary. Then a randomly chosen article sentence(except positive samples) is marked as \textbf{negative item sample}. Only one triplet is picked up to represent the subjective style of each sample even though the summary in each sample may have multiple sentences. This is because often all Oracle-Summary sentence pairs in each article-summary sample have the same styles. Following the extraction of "user, positive item, negative item", this triplet is encoded by GCN. Following the setting of~\cite{kipf2017semi-supervised}, a three-layer GCN is built to obtain node embeddings and a mean-pooling for extracting graph embeddings:
\begin{gather}
  h_{i+1,x} = ReLU  ( M^{-\frac{1}{2}} \hat{A} M^{-\frac{1}{2}} h_{i,x} W) \\
  h_{0,x} = X \\
  h_x = MeanPooling(h_{3,x})
\end{gather}
where $X \in R^{N \times D}$ is the input node embedding matrix of SynGraph and $\hat{A} = A + I$ is the adjacent matrix of SynGraph with self loop. The degree matrix is denoted by $M_{ii} = \sum _j \hat{A}_{ij}$. Then embeddings of the triplet $h_{user},h_{pos},h_{neg}$ are used to calculate the margin triplet loss:
\begin{gather}
  L = max(0,score_{neg}-score_{pos} + \gamma) \label{equ:score} \\ 
  where \ score_{x} = \sigma (W[h_{x};h_{user}]+b)
\end{gather}
where $\gamma$ is the margin and $h_x$ is the graph embedding of x(user/pos/neg). After the model is trained, the graph embeddings of the user and pos $h_{user} \in R^D, h_{pos} \in R^D$ can be obtained and then concatenated as the subjective style embeddings $h_{subj} = [h_{user};h_{pos}] \in R^{2*D}$.

A similar model for extracting features from the syntactic structure is TreeLSTM~\cite{tai2015improved} which operates on the original dependency parse tree. Compared to TreeLSTM, our GCN-based model focuses more on the syntactic structure instead of semantics and it uses the LTRS task to fit the scenario of abstractive summarization. Another benefit of this approach is that there is no trigger order and the computing over all nodes is fully parallelized. 

For comparision Graph2vec~\cite{narayanan2017Graph2vec:} is tested as a baseline for extracting graph embedding in unsupervised fashion. Motivated by neural document embedding models, Graph2vec only considers the structure of the graph and takes subgraphs as "words" to embed graph features. 
\subsection{Cluster, Divide and Train}
We cluster samples with similar styles so we can obtain dataset with single style in unsupervised way. Kmeans++~\cite{arthur2007k-means++:} is a proper method to cluster the subjective style embeddings of all samples in this experiment. The $k$ selection is based on the visualization results of subjective style embeddings via tsne~\cite{dermaaten2008visualizing}. The training set is then divided based on the clustering results. If the subjective style is correctly detected, then each part of a dataset should only keep one style, and thus eliminate the impact of subjective bias.

To assess the impact of different style, three representative models are assessed: Pointer Generator\cite{see2017get}, a classical Recurrent Neural Network(RNN) based seq2seq model with copy mechanism; Transformer~\cite{vaswani2017attention}, which has been the mainstream architecture in the seq2seq task; T5~\cite{raffel2019exploring} which stands for Pretrain-Then-Finetune style for natural language processing. Each style-oriented model is trained on divided per style datasets and evaluated on the full test set.

\section{Experimental Settings}
\subsection{Dataset}
Experiments are conducted on the processed version of the CNN-DM dataset~\cite{see2017get} which contains 287227 samples for training, 13368 for validation and 11490 for testing with 685.2 words for an article and 52 words for a summary on average. Some very short articles that can not obtain triplets are excluded with 280000 samples remained for training.
\subsection{LTRS}
The hidden size of node embedding is set to 256. Each graph contains nodes at most twice the number of sentence length because dependency nodes are included. Each sample contains three graphs. Since the model size is small the batch size can be set to 2048 on a GPU with 8G memories. The optimizer for training is a standard Adam optimizer with default hyperparameters. The GCN model only contains a learnable parameter amount of 90k and takes less than two minutes to train an epoch on a GTX 2070. The margin $\gamma$ in triplet loss is set to 0.5.
\subsection{Cluster and Divide}
Based on tsne's visualization(which is shown in Figure \ref{fig:clustering_result}) result the $k$ is set to 4 for clustering and then the training set is divided into corresponding 4 parts. For each cluster, only the top 45000 samples(because the smallest cluster only contains about 45000 samples) closest to the cluster centroid were picked up to formulate the divided dataset. Three baseline dataset division methods are built for comparision(each divided dataset contains 45000 samples so that the results are comparable).

\begin{itemize}
\setlength{\itemsep}{0pt}
\setlength{\parsep}{0pt}
\setlength{\parskip}{0pt}
\item \textbf{Baseline 0}: Obtain top 11250 samples closest to corresponding centroid from 4 clusters(4*11250). This baseline is set for observing the impact of maximum bias(equal amount for each style, no preference).
\item \textbf{Baseline 1}: Obtain 45000 samples from four clusters based on their original percentages(27.95\%, 32.89\%, 22.93\%, 16.22\%) in the whole dataset, which are 12578, 14801, 10320, 7299 samples for the four clusters respectively. This baseline actually scales down the whole training dataset to 45000 samples.
\item \textbf{Baseline 2}: Same as Baseline 0 but pick up 11250 samples that are furthest from each cluster centroid. This baseline considers those samples that don't have an obvious single style.
\end{itemize}
\subsection{Abstractive Summarization model}
Three representative abstractive summarization models are tested in our experiments which are Pointer Generator Network(PGNet), Transformer, and T5. PGNet is a strong baseline which utilizes copy mechanism to improve the readability of summary but also brings the copy-too-much problem. Transformer has been the new paradigm of seq2seq modeling but it is more data-hungry than RNN based models. Based on the setting of the translation task, the model uses an encoder and a decoder with $N=12$ self-attention blocks and $h=16$ masked self-attention heads in each block. Both PGNet and Transformer are trained until the training and validation loss didn't improve anymore.

The small version of T5 which contains about 60 million parameters is finetuned for 10 epochs on each dataset. Although the pretraining process of T5 contains a multi-task target including summarizing on CNN-DM dataset, the impact of this task is small compared to the Language Modeling task on C4 corpus~\cite{raffel2019exploring}. The pretrained model can been seen as a model having the text-to-text summarization ability instead of a fully finetuned summarization model. Hence the results of further finetuning can still be used to evaluate the style's influence.
\subsection{Metrics}
ROUGE-1, ROUGE-2, ROUGE-L~\cite{lin2004rouge:} and METEOR(MTR)~\cite{banerjee2005meteor:} are N-gram-based metrics for summarization. GLEU~\cite{wu2016google's} is a strict version of BLEU which is the standard metric of machine translation. BERT Based Score(BERT)~\cite{zhang2019bertscore:} is an embedding based metric scoring the semantic similarity between gold and generated summaries. These six metrics give a multi-view evaluation on the similarity between generated summaries and gold summaries which partially reflects the correctness, readability and fact-consistence of the model generated summaries. But these "Summary Related" metrics can not measure the abstractness of summaries since copied(extracted) summaries can also achieve high scores under these metrics. So three "Article Related" metrics are introduced: Novel Unigram Ratio(N1), Novel Bigram Ratio(N2), Average Jaccard Similarity(JS) to describe the extent that summaries are more like "copied" or "generated". Novel metrics measure the proportion of new words in summaries compared to articles. For each summary sentence, the Jaccard Similarity is calculated between it and its corresponding oracle sentence in the article, which also reflects the abstractness. Finally, Oracle Hit is calculated to evaluate the Natural Language Understanding ability of summarization models. Oracle Hit is defined as the precision that generated summary shares the same oracle sentence with gold summary, which means generated summaries "hit" the key content of the article.

For Summary Related metrics, the higher the better. For Article Related metrics, the closer to gold summaries' score the better(Neither too extractive nor too abstractive). The higher the Oracle Hit, the better model performs. 
\section{Results and Analysis}
In this section, we first review the clustering results, and then give a detailed analysis of each cluster. Finally, comprehensive summarization results are reported and analysed on all models and all datasets.
\subsection{Clustering}
\begin{figure}[htbp]
  \centering
  \includegraphics[scale=0.5]{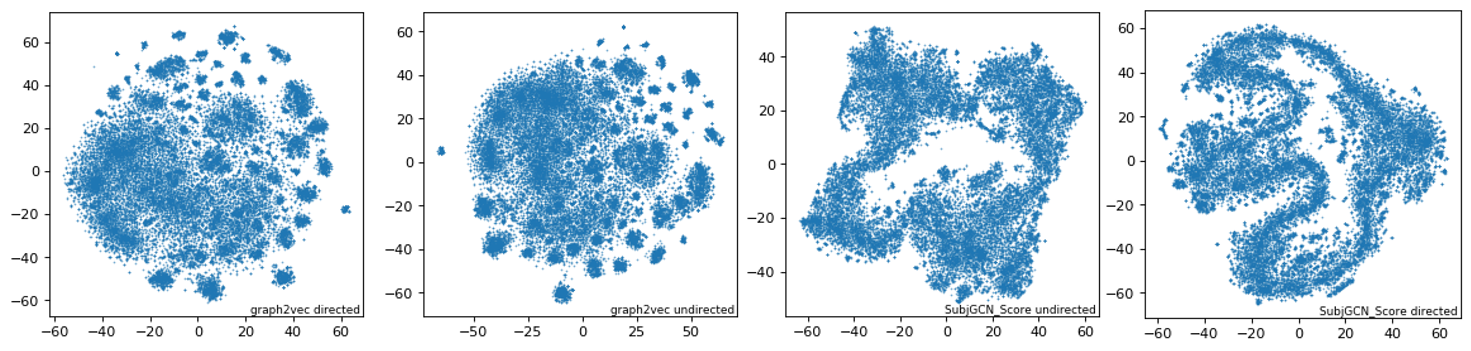}
  \caption{Visualization of subjective style embeddings using different models on SynGraph.}
  \label{fig:clustering_result}
\end{figure}

The subjective embeddings from four settings: [Graph2vec, SubjGCN] * [undirected SynGraph, directed SynGraph] are visualized in Figure \ref{fig:clustering_result}. The SubjGCN means GCN in LTRS task for extracting Subjective Style representation with linear Score layer. It shows that Graph2vec extracts too many fine-grained features leading to many small groups. It may only group samples with almost the same sentence structure and ignore the structure relationship between Summary and Oracle. The SubjGCN successfully catches graph features at a proper level and divides the dataset into obvious four groups. Based on the clustering quality the SubjGCN with undirected SynGraph is chosen as the final result for clustering. Kmeans++ allocates 78292, 92045, 64199, 45464 samples for each group respectively. In order to make the results comparable, all clusters and baseline datasets have 45000 samples as mentioned before.
\subsection{Inside Each Style}
\begin{figure}[htbp]
  \centering
  \includegraphics[scale=0.3]{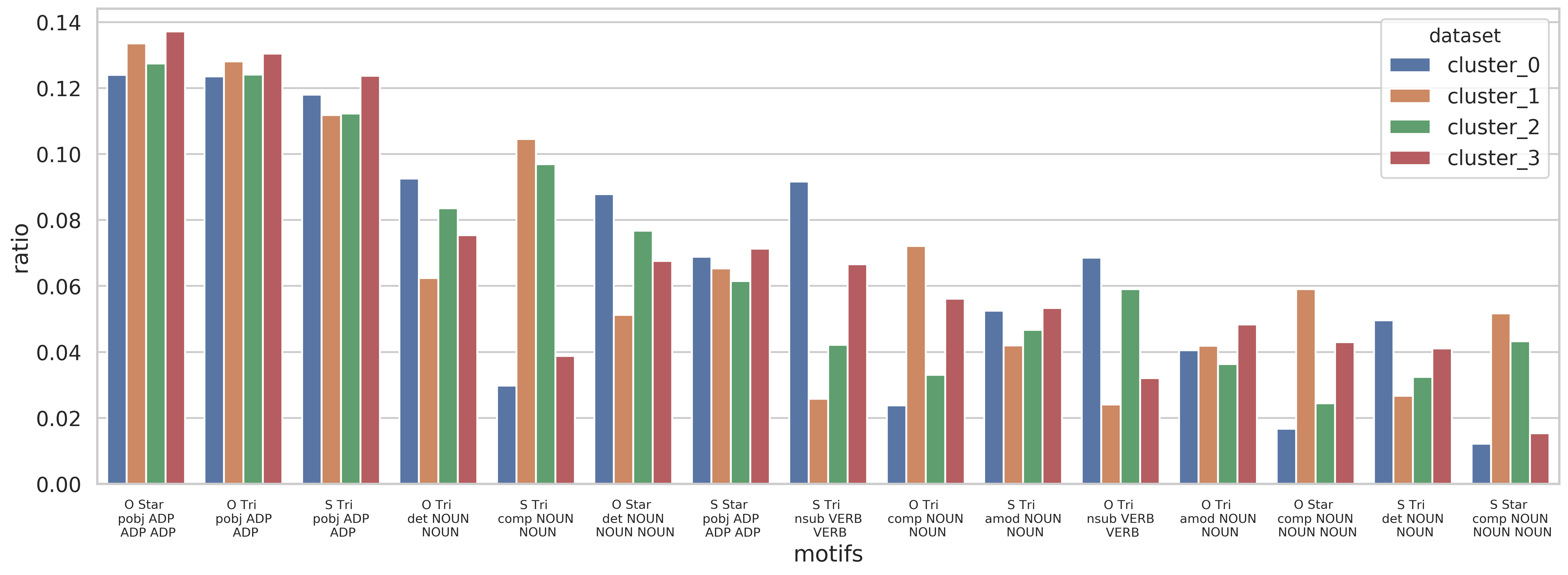}
  \caption{top 15 graph motifs ratio of each style.}
  \label{fig:motifs}
\end{figure}

First, some motifs in graphs are counted to discover what are the featured styles or graph structures inside each cluster. Three representative motifs were chosen which are three-braches star(Star), triangle path(Tri) and Four-step path(Four). For example, "O Star pobj ADP ADP ADP" means there exists a three-branches star motif with pobj(object of a preposition) node as the center which connects three ADP(adposition) nodes in an Oracle sentence. There are a total of 43878 motifs in the whole datasets which present a long-tailed distribution. Complication derived from numerous motifs(that come from the three kinds of shapes alone) also indicates that using graph structures only to embed subjective style is impracticable. Top 15 motifs and visualization of their distribution in each cluster can be compared in Figure \ref{fig:motifs}. The ratio means the average proportion on a graph(e.g. 0.1 means this type of motifs accounts for an average of 10\% in each graph). As for the top 3 motifs, their distributions in four clusters do not make a significant difference and ratios all surpass 0.1. So these three motifs present most in sentence structures but are not featured in each cluster. Cluster 1 and cluster 2 tend to take a bigger proportion of "S Tri comp NOUN NOUN" which is a triangle motif with a complement connecting two noun words in Summary. Cluster 0 prefers "S Tri nsub VERB VERB" which is a triangle motif with a nominal subject connecting two verbs. It is noticeable that all motifs are extracted in SynGraph in which the dependency nodes are merged. So the "nominal subject connecting two verbs" means there are two verbs in this sentence that have nsubj dependency relationship with other words. These different degrees of preference indicate that our LTRS task do group samples with some certain sentence structures together. But the content of the style is still vague.
\definecolor{root}{RGB}{215,47,16}
\definecolor{co-occurrence}{RGB}{154,205,50}
\begin{table}[htbp]
  \footnotesize
  \centering
  \begin{tabular}{|c|l|}
  \hline
  Samples                   & \multicolumn{1}{c|}{Summary-Oracle Pair}                                             \\ \hline
  \multirow{2}{*}{Sample 1} & non-christians \textcolor{co-occurrence}{will} not \textcolor{root}{go} see 'son of god' because it 's a terrible movie.          \\ \cline{2-2} 
                            & \textcolor{co-occurrence}{will} americans \textcolor{root}{embrace} hollywood version of noah story? probably not. \\ \hline
  \multirow{2}{*}{Sample 2} &
    \begin{tabular}[c]{@{}l@{}}strident ross \textcolor{co-occurrence}{longhurst} \textcolor{root}{wielded} a loudhailer outside \textcolor{co-occurrence}{court} before he was jailed for 28  days for not paying up.\end{tabular} \\ \cline{2-2} 
                            & former university lecturer ross \textcolor{co-occurrence}{longhurst} \textcolor{root}{used} a loudhailer outside \textcolor{co-occurrence}{court}.           \\ \hline
  \multirow{2}{*}{Sample 3} &
    \begin{tabular}[c]{@{}l@{}}former flame former playboy model miss \textcolor{co-occurrence}{becirovic} \textcolor{root}{dated} al pacino in 1972 when he was making the godfather.\end{tabular} \\ \cline{2-2} 
                            & diana \textcolor{co-occurrence}{becirovic} \textcolor{root}{dated} the actor in 1972.\\ \hline
  \end{tabular}
  \caption{Samples in training set from cluster 1. Upper one is Oracle and below is Summary. A red denotes the root predicate in the SynGraph and a green one denotes the co-occurrent noun/verb.}
  \label{table:co-occurrence}
  \end{table}

  Motifs of Oracle or Summary only present features for a single sentence not Summary-Oracle pair and the joint distribution of Oracle and Summary motifs will be sparse which is hard to analyze. Hence from the perspective of transformation, we visualize the dependency trees and annotate the co-occurrence of nouns and verbs to formulate a Summary-Oracle graph. As can be seen in Figure \ref{fig:clustering_example}, nodes are POS and edges are dependency relationships. The green edge stands for dependency relationship in Oracle sentence and the blue one for Summary sentence. The orange edge annotates the co-occurrence of nouns and verbs between Oracle and Summary. Node with self-loop is the root node in the dependency parsing results. Several statistics(counts of POS in Summary and Oracle, ratios of three kinds of edges) reveal the main features of each cluster:
\begin{figure}[htbp]
  \centering
  \subfigure[Cluster 0]{
  \begin{minipage}[t]{0.23\linewidth}
    \centering
    \includegraphics[scale=0.04]{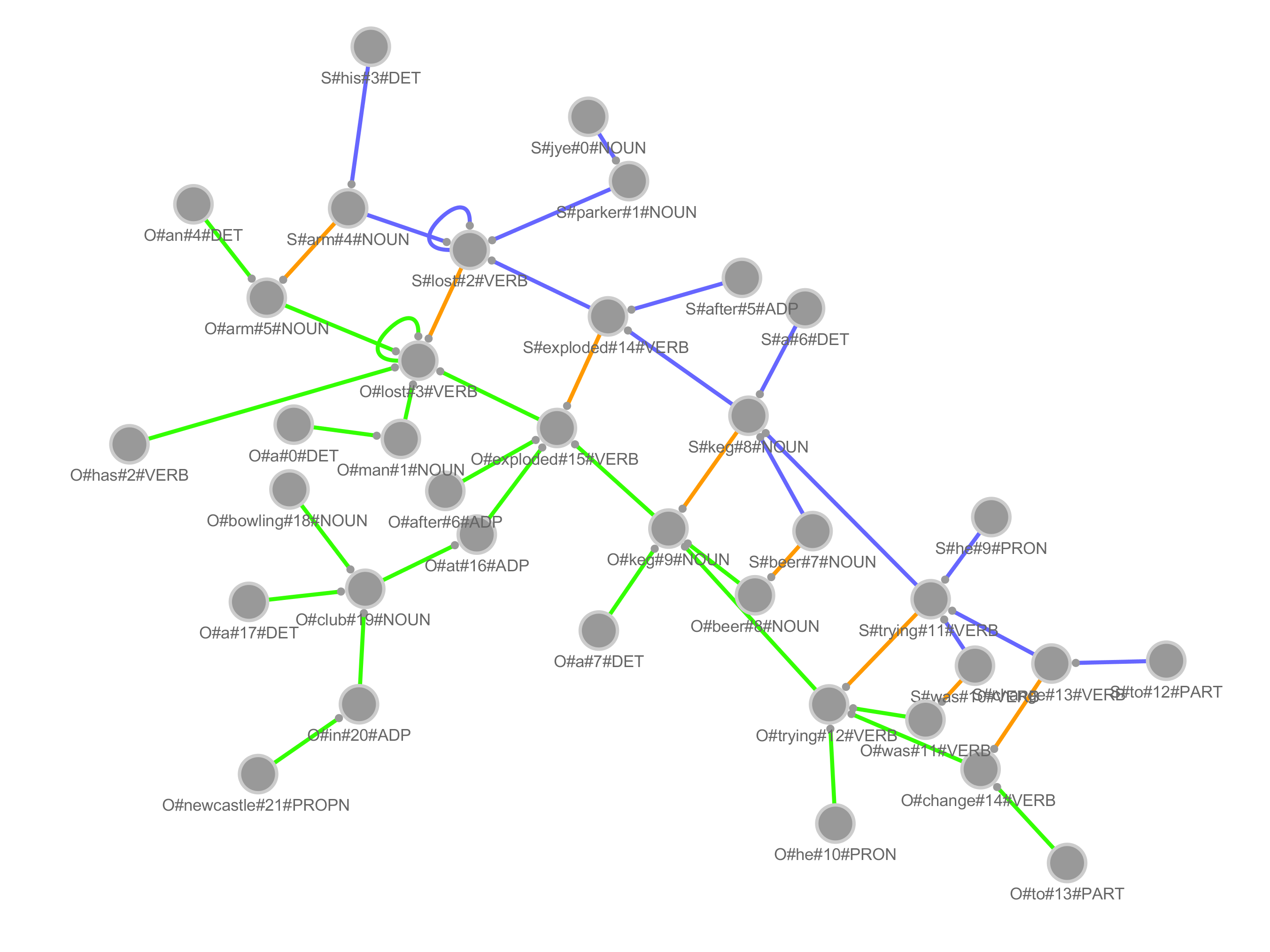}
    \includegraphics[scale=0.04]{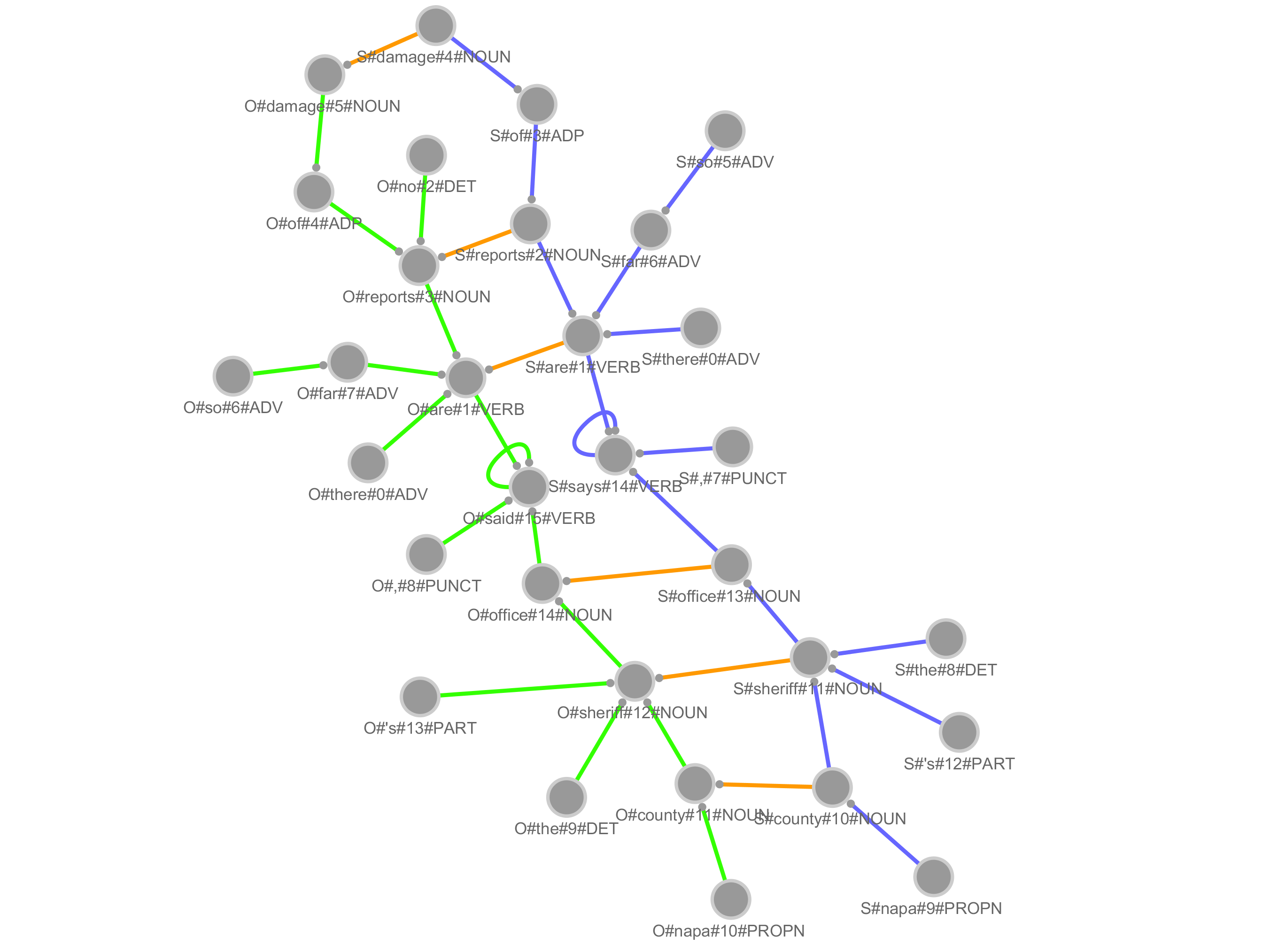}
    \includegraphics[scale=0.04]{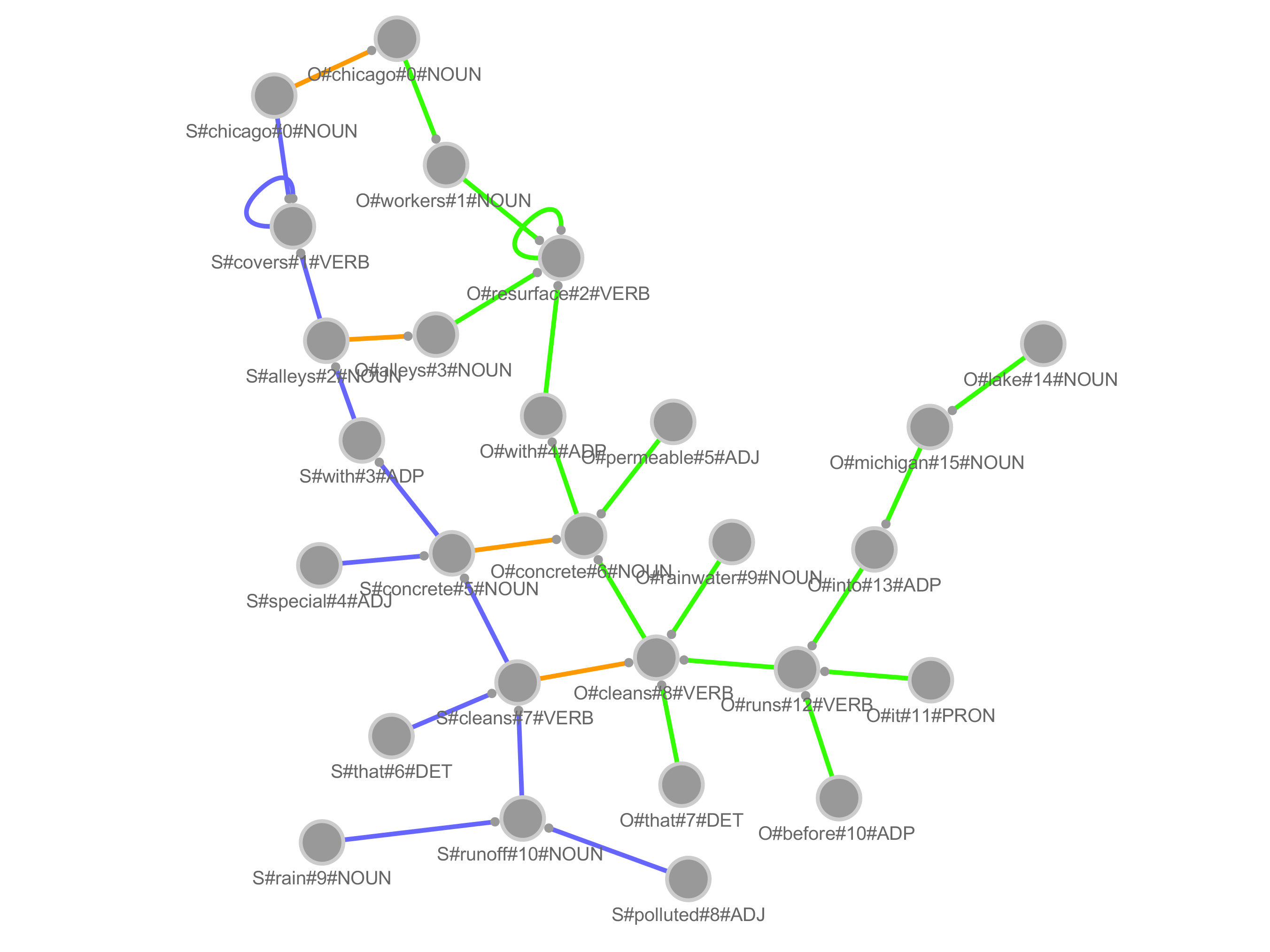}
    \label{fig:c0}
  \end{minipage}%
  }
  \subfigure[Cluster 1]{
  \begin{minipage}[t]{0.23\linewidth}
    \centering
    \includegraphics[scale=0.04]{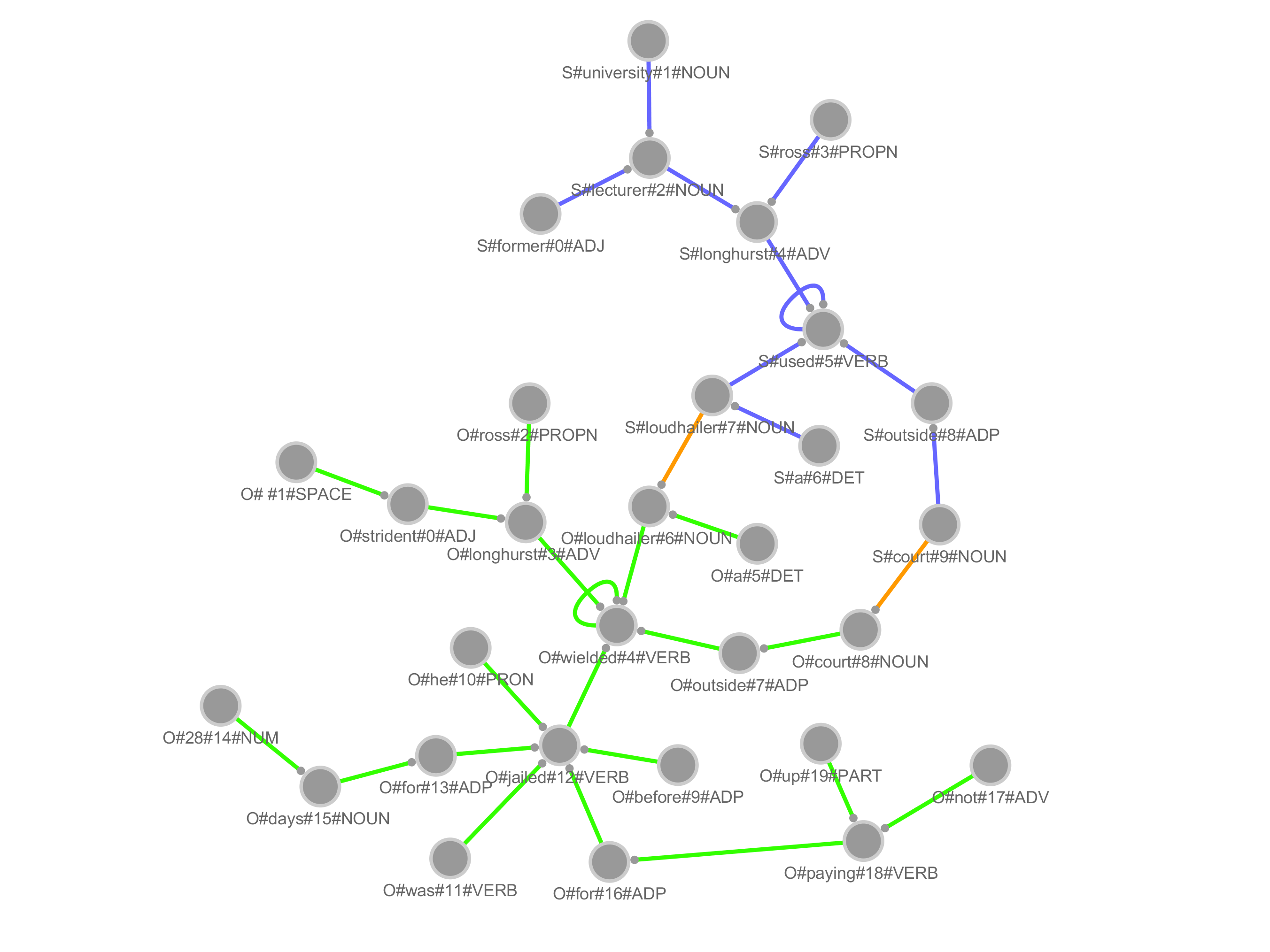}
    \includegraphics[scale=0.04]{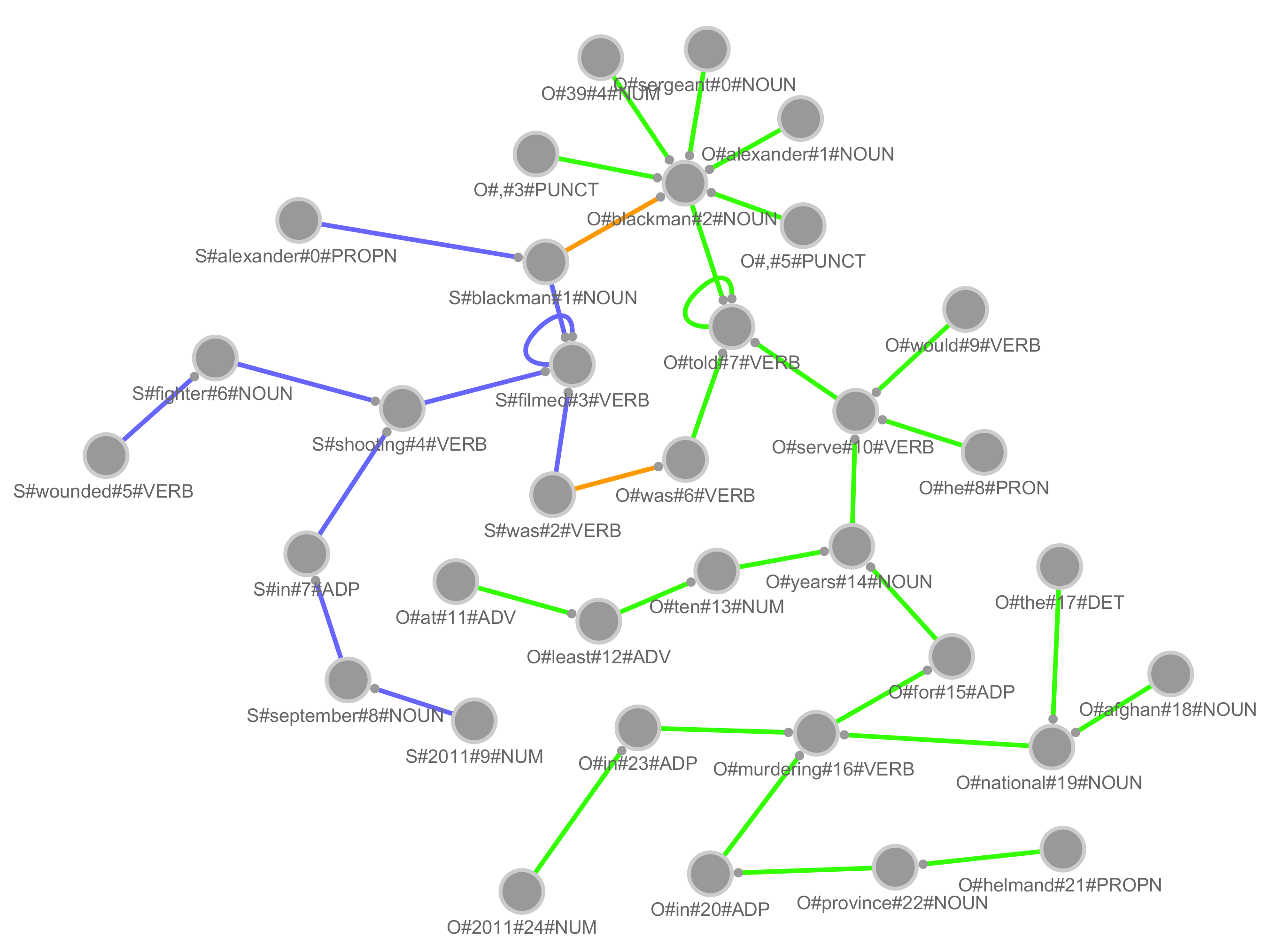}
    \includegraphics[scale=0.04]{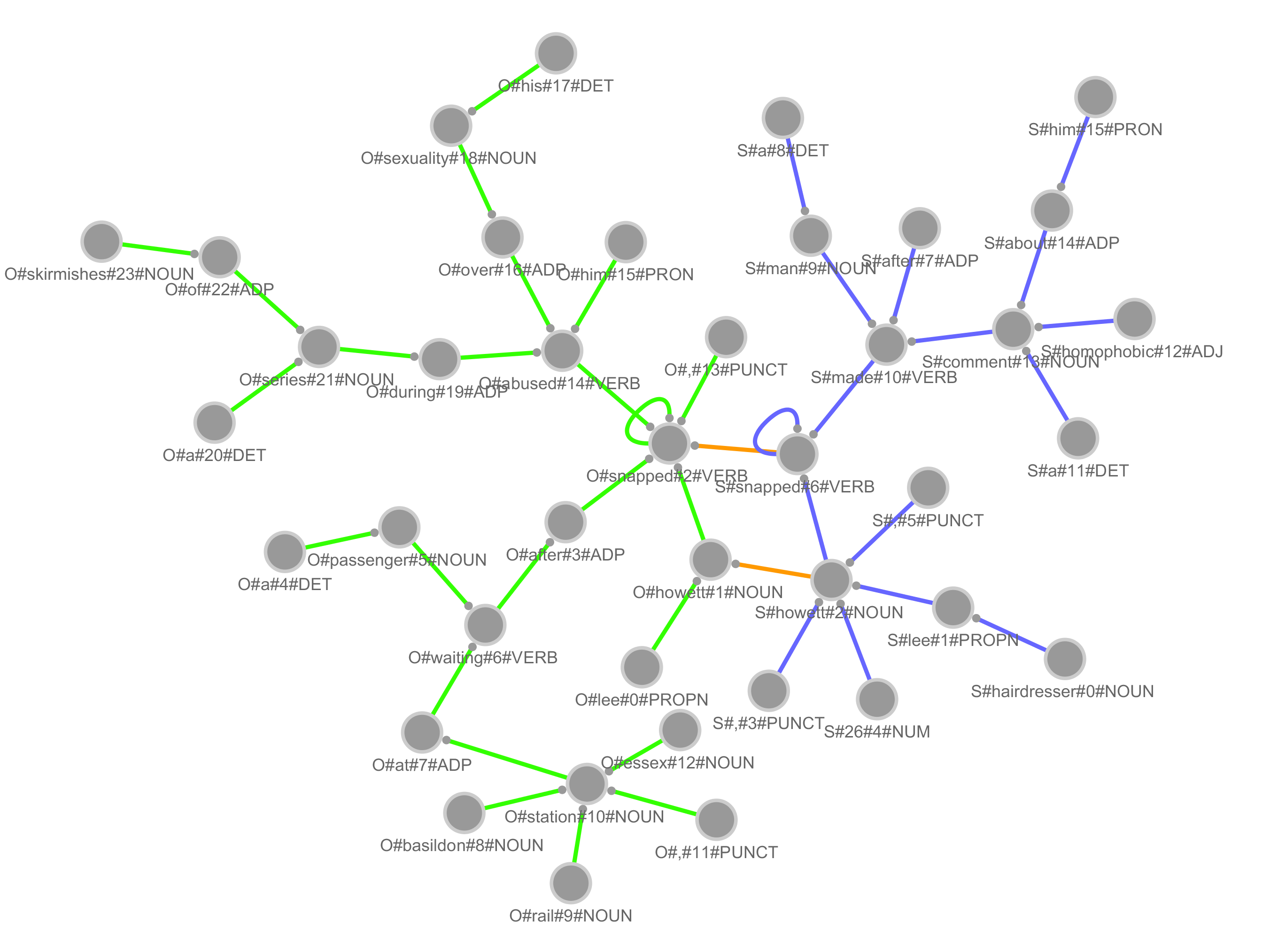}
    \label{fig:c1}
  \end{minipage}%
  }
  \subfigure[Cluster 2]{
  \begin{minipage}[t]{0.23\linewidth}
    \centering
    \includegraphics[scale=0.04]{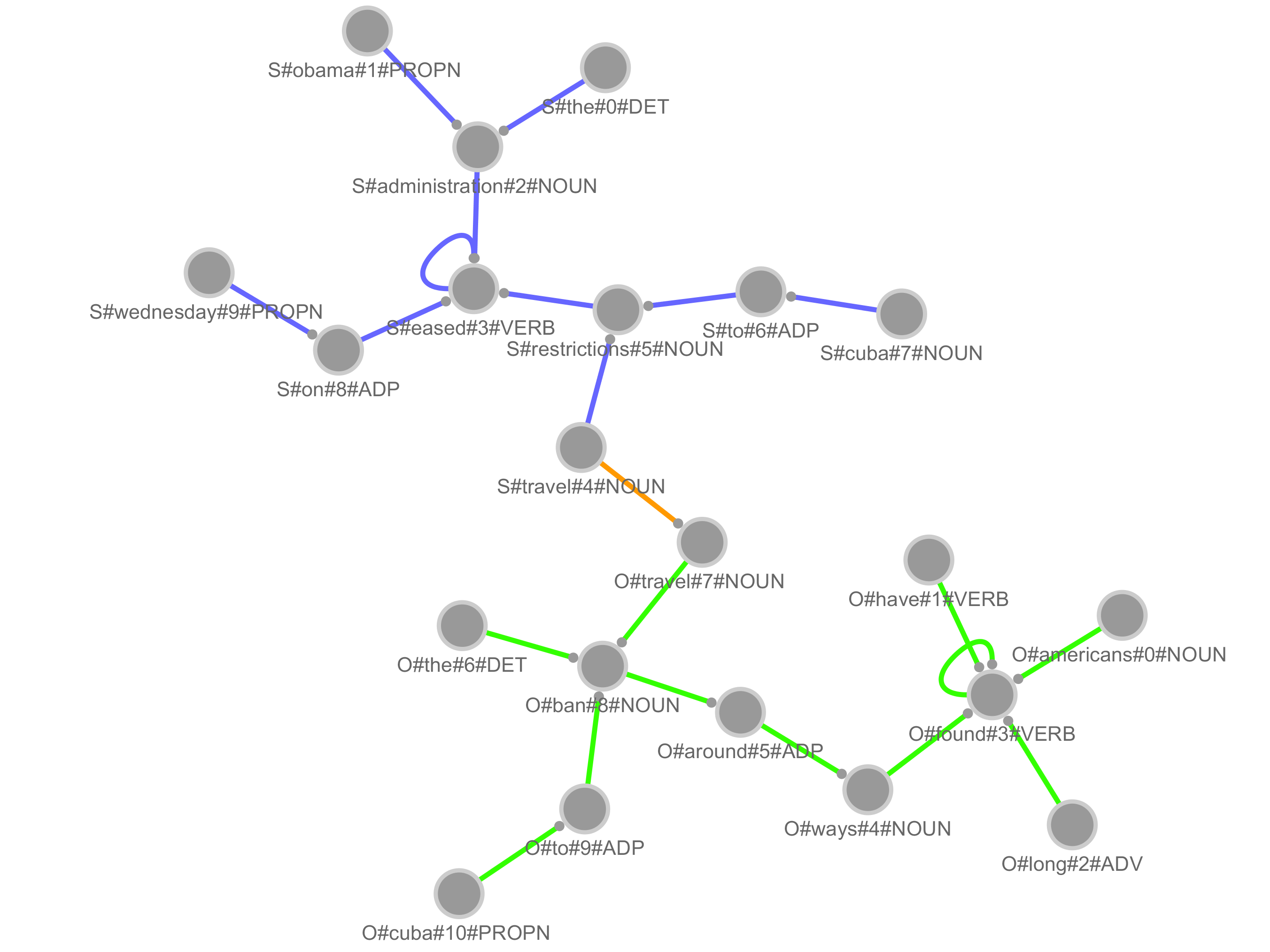}
    \includegraphics[scale=0.04]{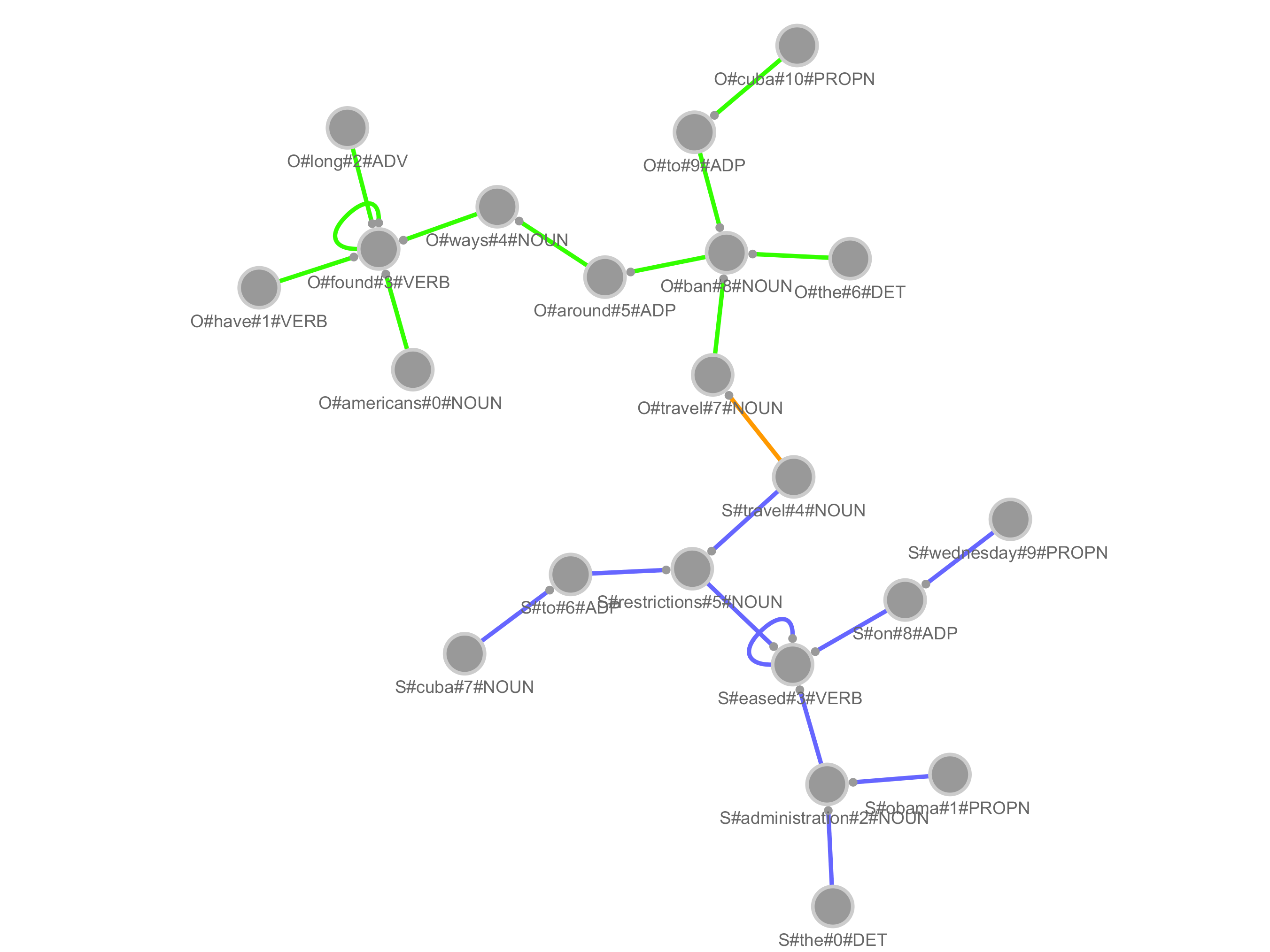}
    \includegraphics[scale=0.04]{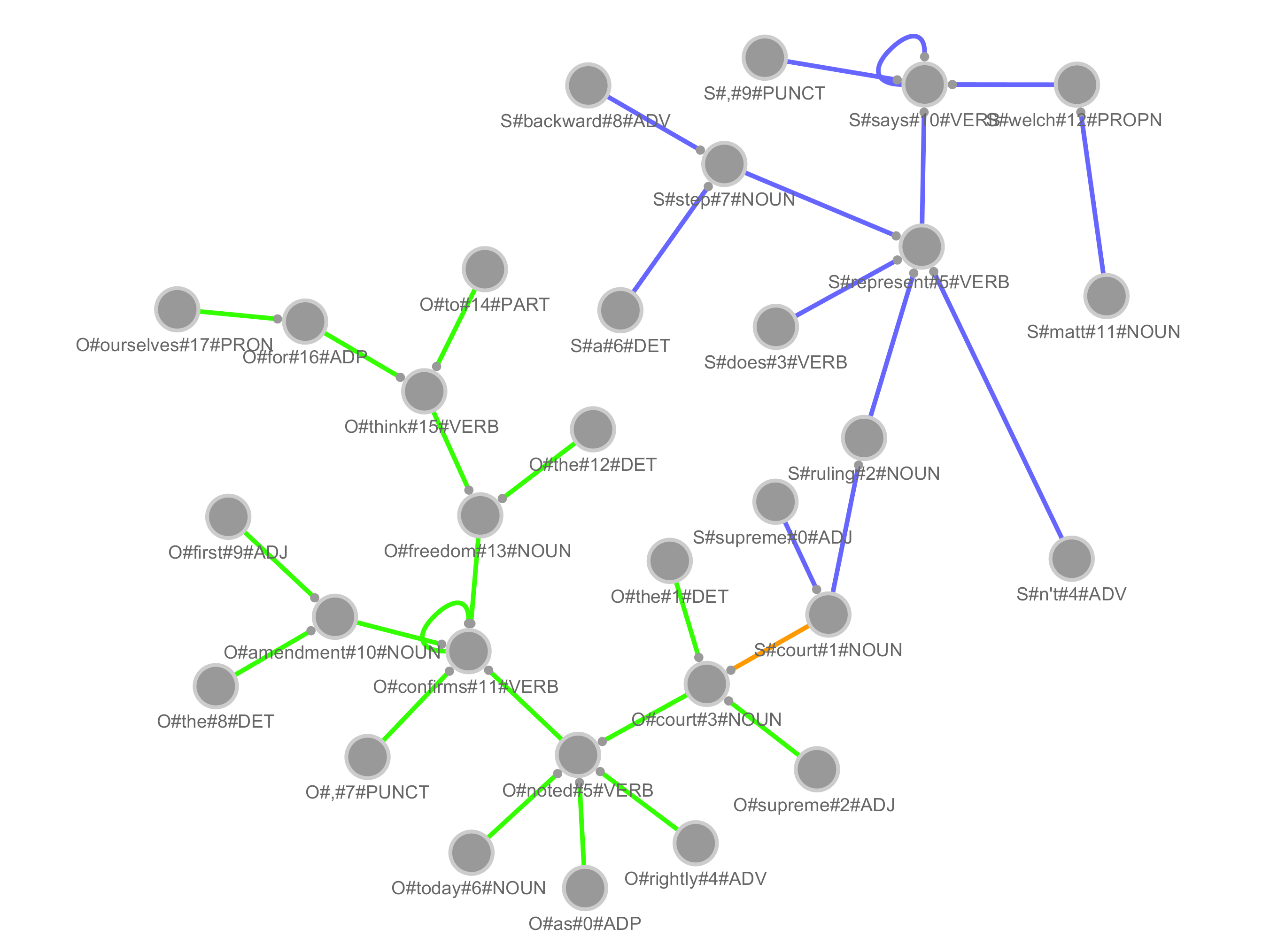}
    \label{fig:c2}
  \end{minipage}%
  }
  \subfigure[Cluster 3]{
  \begin{minipage}[t]{0.23\linewidth}
    \centering
    \includegraphics[scale=0.04]{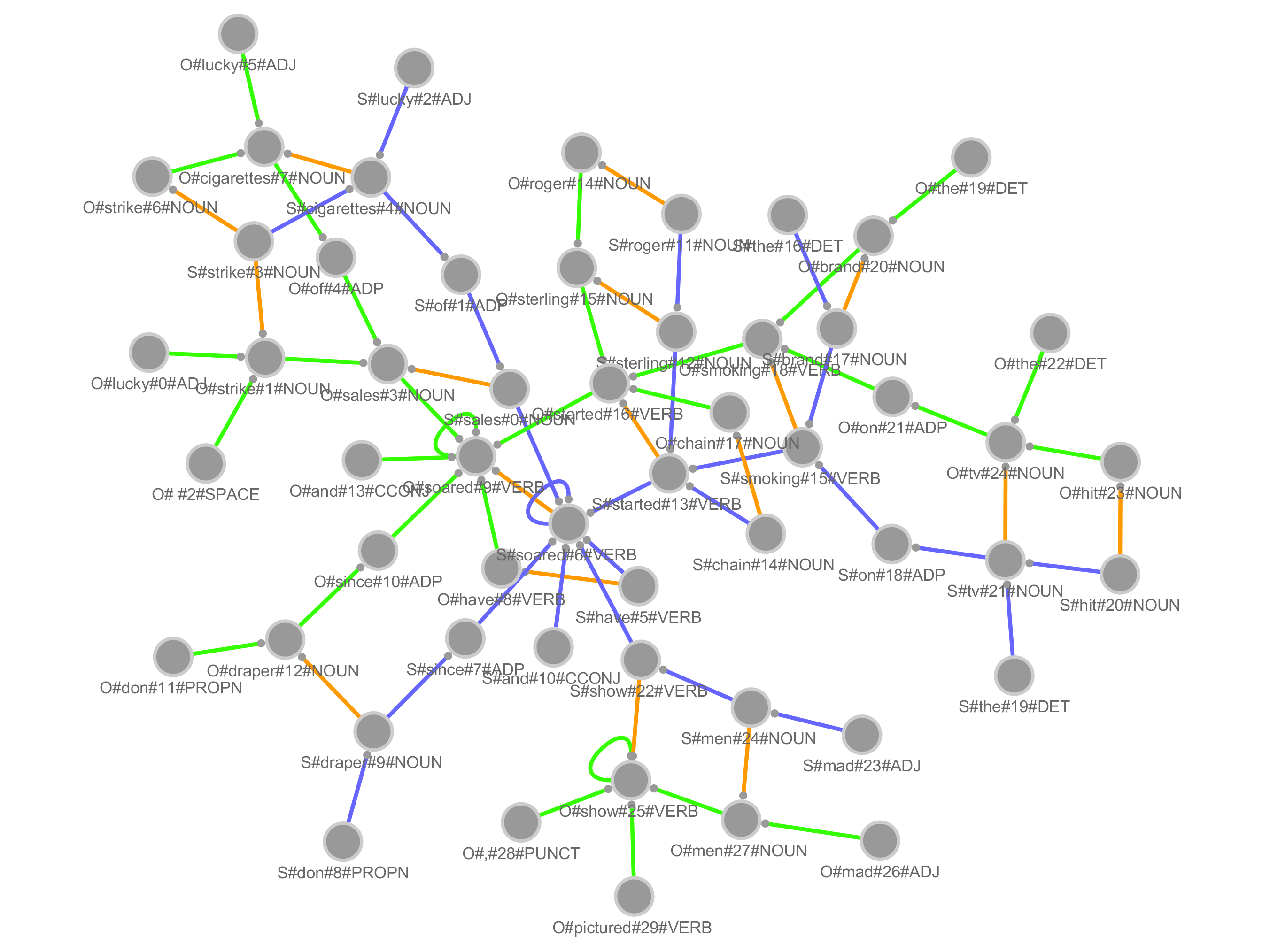}
    \includegraphics[scale=0.04]{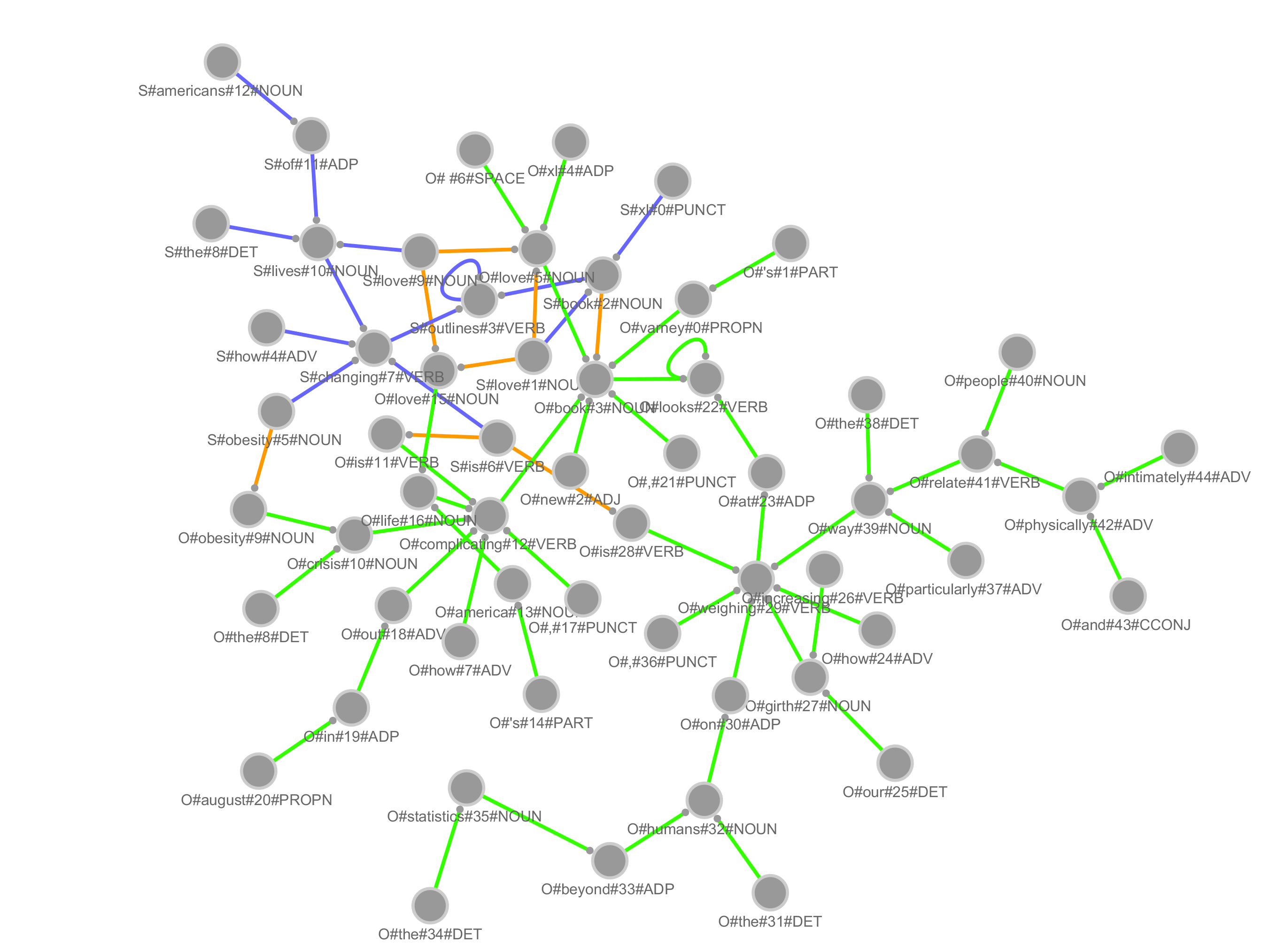}
    \includegraphics[scale=0.04]{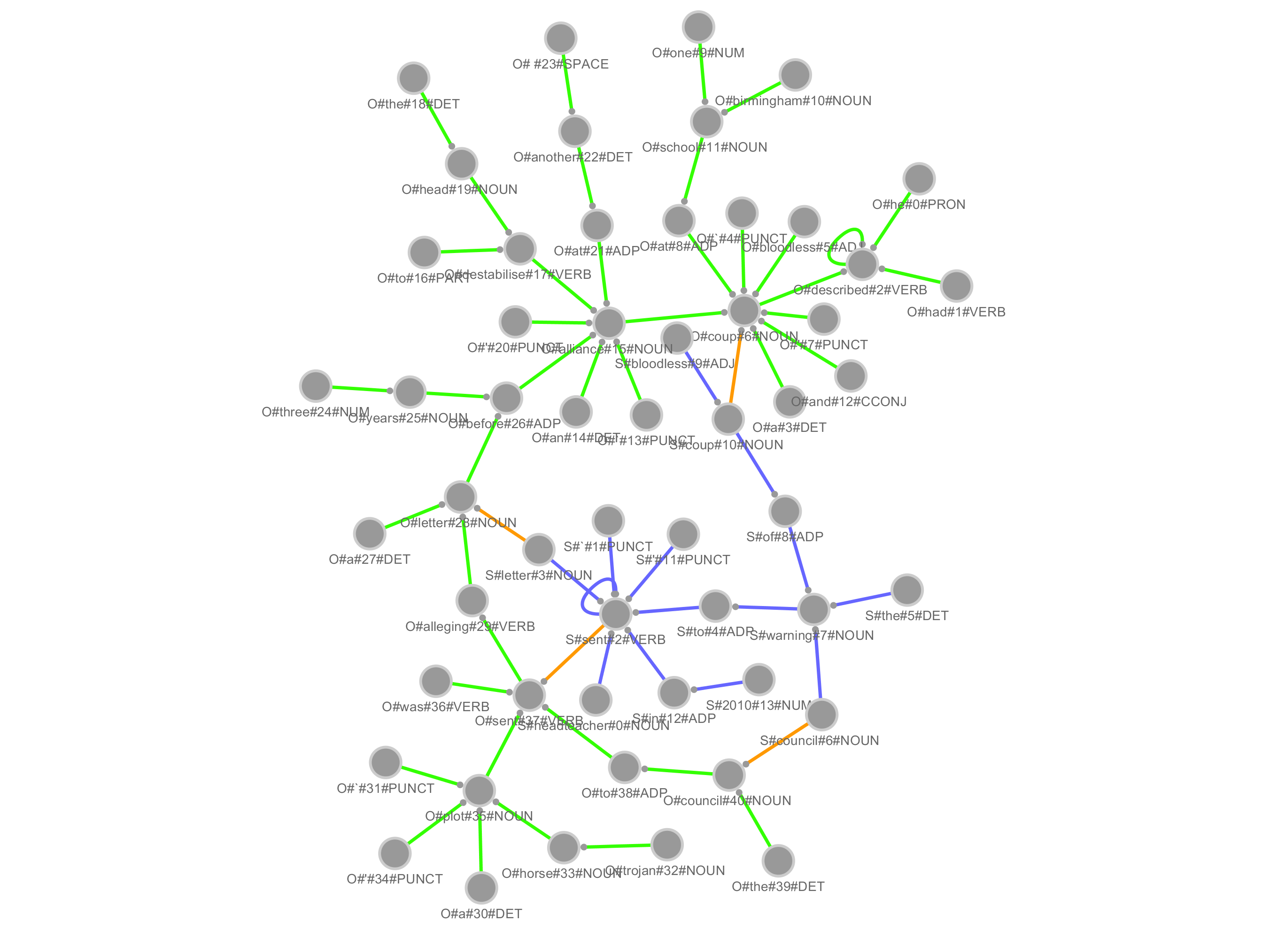}
    \label{fig:c3}
  \end{minipage}%
  }
  \caption{Summary-Oracle graphs from four clusters.}
  \label{fig:clustering_example}
\end{figure}

\begin{itemize}
  \setlength{\itemsep}{0pt}
  \setlength{\parsep}{0pt}
  \setlength{\parskip}{0pt}
  \item \textbf{Cluster 0}: Oracle and Summary have almost the same amount of nodes. There are two main lines in two sentences' graph that contain the main content. Two lines are almost completely aligned with the orange edge on every node. This kind of Summary-Oracle graph represents sentence rewritting to a very small extent. Human copied the main part of the sentence and compressed some adjuncts. So the lengths of Oracle and Summary are close and there are many alignments in graphs of this style.
  \item \textbf{Cluster 1}: In this style there often exist two or more alignments near the root node. Usually, this pattern indicates that human transformed(or kept) the root predicate of a sentence, reserving some context of this predicate then changed a lot on the other parts.
  \item \textbf{Cluster 2}: Summary-Oracle graphs in cluster 2 usually have at most one or two co-occurrence and the distance between two roots is far. Under this circumstance, we assume that a human summary was formulated from multiple Oracle sentences, and therefore, summary graphs are hardly aligned with oracle graphs.
  \item \textbf{Cluster 3 }: Same as cluster 0, graphs in cluster 3 have many alignments but the length of Oracle sentence is much longer than that of sentence in cluster 0. The average compression ratio is also the highest among the four clusters. This style corresponds to summarizing long sentences. 
  \end{itemize} 

  The Summary-Oracle visualization results can give a quick review of each style. In Table \ref{table:co-occurrence} there are three examples from cluster 1. For each sample, human compresses the Oracle(up) to Summary(down). Red words are root predicates that define the action in the sentence content and human usually keep this word or keep this meaning in style 1. Green words(co-occurrent context words) are also kept in style 1. Thus this is a classical compression method: choose an Oracle sentence, keep its predicate and supply some other information from related sentences, then compress the rest modifiers. In example 2, "longhurst wielded a loudhailer outside court" is the fact that obtained from Oracle and human deletes the "before he was jailed for ....", adds information from other sentences("former university lecturer"). The retention of context("longhurst" and "court") locates the predicate that human wants to keep. There may be some noise during building graphs (such as the co-occurrence of "loudhailer" was not recognized because of the wrong POS tagging) but the entire style can still be caught and clustered.
\subsection{Impact of Style}
\begin{table}[htbp]
  \footnotesize
  \centering
  \begin{tabular}{|c|c|c|c|c|c|c|c|c|c|c|c|}
  \hline
  \multirow{2}{*}{\textbf{Model}} &
    \multirow{2}{*}{\textbf{Dataset}} &
    \multicolumn{6}{c|}{\textbf{Summary Related}} &
    \multicolumn{3}{c|}{\textbf{Article Related}} &
    \textbf{NLU} \\ \cline{3-12} 
   &
     &
    \textbf{R1} &
    \textbf{R2} &
    \textbf{RL} &
    \textbf{MTR} &
    \textbf{GLEU} &
    \textbf{BERT} &
    \textbf{N1} &
    \textbf{N2} &
    \textbf{JS} &
    \textbf{Oracle Hit} \\ \hline
  \multirow{1}{*}{\textbf{Gold}}      & \textbf{--}  & --  & --    & --    & --   & --    & -- & 18.9 & 55.7 & 31.2 & -- \\ \hline 
  \multirow{4}{*}{\textbf{PGNet}}      & \textbf{Cluster\_0}  & 32.7   & 12.5    & 22.8    & 26.3   & 13.0    & 85.5 & 2.1 & 7.9 & 67.4 & 36.9 \\ \cline{2-12} 
                                        & \textbf{Cluster\_1} & 34.4    & 14.1   & 24.3    & 27.2    & 14.0   & 86.0    & 3.0 & 10.7 & 60.0 & 39.5 \\ \cline{2-12} 
                                        & \textbf{Cluster\_2} & 32.4    & 12.6   & 23.1    & 25.5    & 13.0   & 85.5    & 2.5 & 10.6 & 57.0 & 37.3 \\ \cline{2-12} 
                                        & \textbf{Cluster\_3} & 33.3    & 13.2   & 23.6    & 26.1    & 13.4   & 85.7    & 2.8 & 12.8 & 56.6 & 39.1 \\ \hline
  \multirow{4}{*}{\textbf{Transformer}} & \textbf{Cluster\_0} & 28.3 & 5.6 & 17.5 & 19   & 8.6 & 84.1 & 16.5 & 71.1 & 20.6 & 33.9 \\ \cline{2-12} 
                                        & \textbf{Cluster\_1} & 29.7 & 6.4 & 18.6 & 20.5 & 9.4 & 84.6 & 16.9 & 70.0 & 21.3 & 35.9 \\ \cline{2-12} 
                                        & \textbf{Cluster\_2} & 28.8 & 5.9 & 18   & 19.9 & 9.1 & 84.4 & 17.9 & 71.3 & 20.4 & 34.2 \\ \cline{2-12} 
                                        & \textbf{Cluster\_3} & 29.1 & 6.0   & 18.1 & 19.8 & 9.0   & 84.4 & 16.8 & 70.6 & 21.0 & 34.5  \\ \hline
  \end{tabular}
  \caption{Cluster results of PGNet and Transformer.}
  \label{table:pg_tr}
  \end{table}

First, the results of PGNet and Transformer which both have no pretraining processes are listed in Table \ref{table:pg_tr}. The divided datasets contain only about one-sixth of the whole dataset so the result has a gap from state-of-the-art. But some patterns can be found comparing results among clusters. Whether it is PGNet or Transformer, models trained on cluster 1 obtain the best Summary Related score and Oracle Hit. Then the models trained on cluster 3 ranks the second. Styles in cluster 1 and cluster 3 seem to be better patterns for the model to memorize and generalize. On the contrary, the style to extract(choose and copy) a whole sentence(cluster 0) may be the easiest way for human to summarize document, but not for deep learning models since the models need to "choose" the right sentence first. It can be concluded from the lowest Oracle Hit score of cluster 0. Also, Transformer and PGNet have their preferences. With copy mechanism, PGNet performs nearly the same on cluster 0(copy more) and cluster 2(copy less). But Transfomer results on cluster 0 is worse than those on cluster 2 for all Summary Related metrics. It could also be summarized that there is a significant difference between the PGNet and Transformer results of N1, N2 and JS. PGNet prefers copying(less Novel metrics and higher JS than gold) and on the contrary Transformer prefers generating from scratch.
\begin{figure}[htbp]
  \centering
  \begin{minipage}[t]{0.45\linewidth}
  \centering
  \includegraphics[scale=0.4]{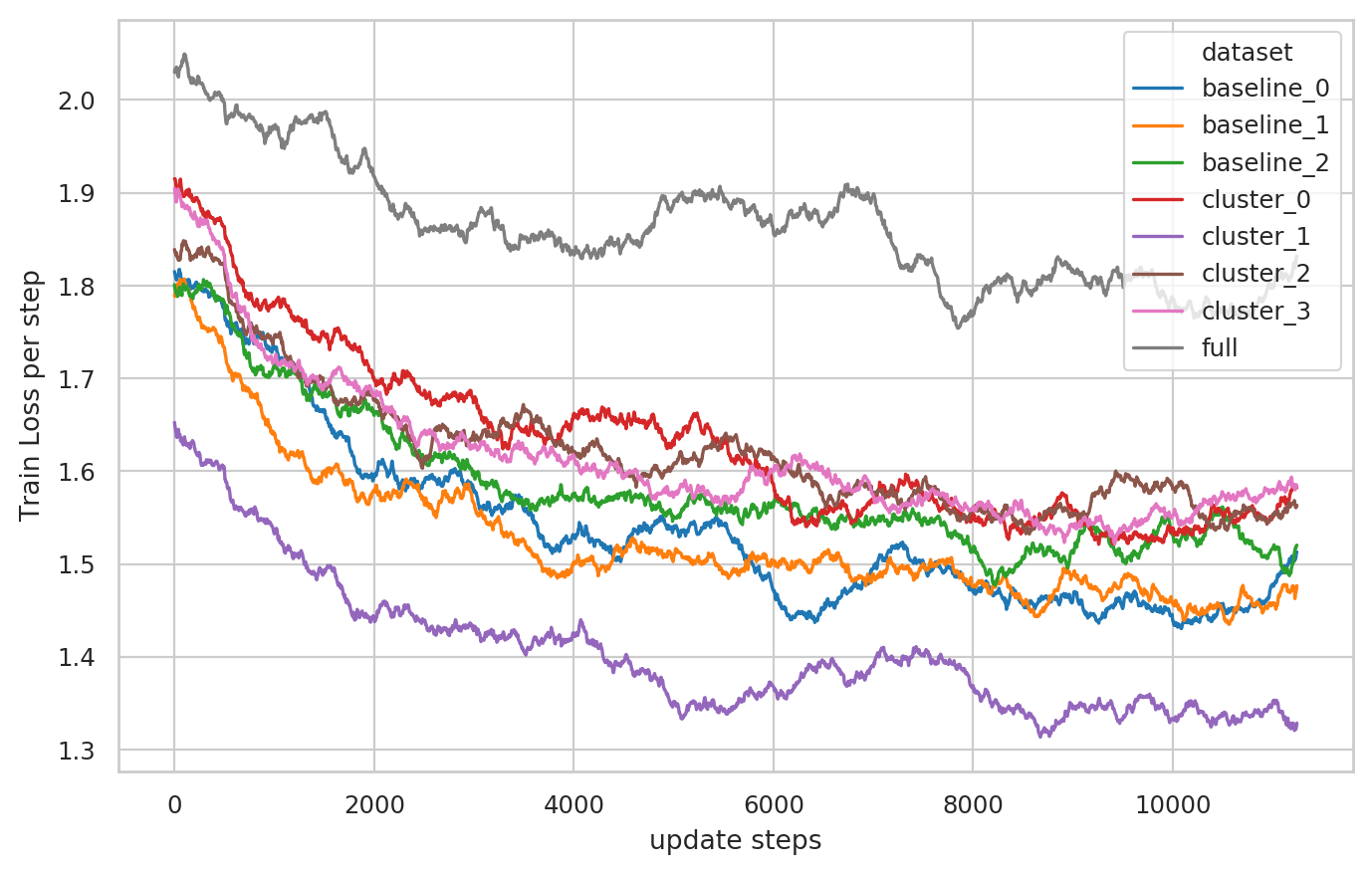} 
  \caption{Fine-tune loss curves on T5-small model.}
  \label{fig:t5_result}
  \end{minipage}%
  \hspace{1mm}
  \begin{minipage}[t]{0.5\linewidth}
  \centering
  \includegraphics[scale=0.55]{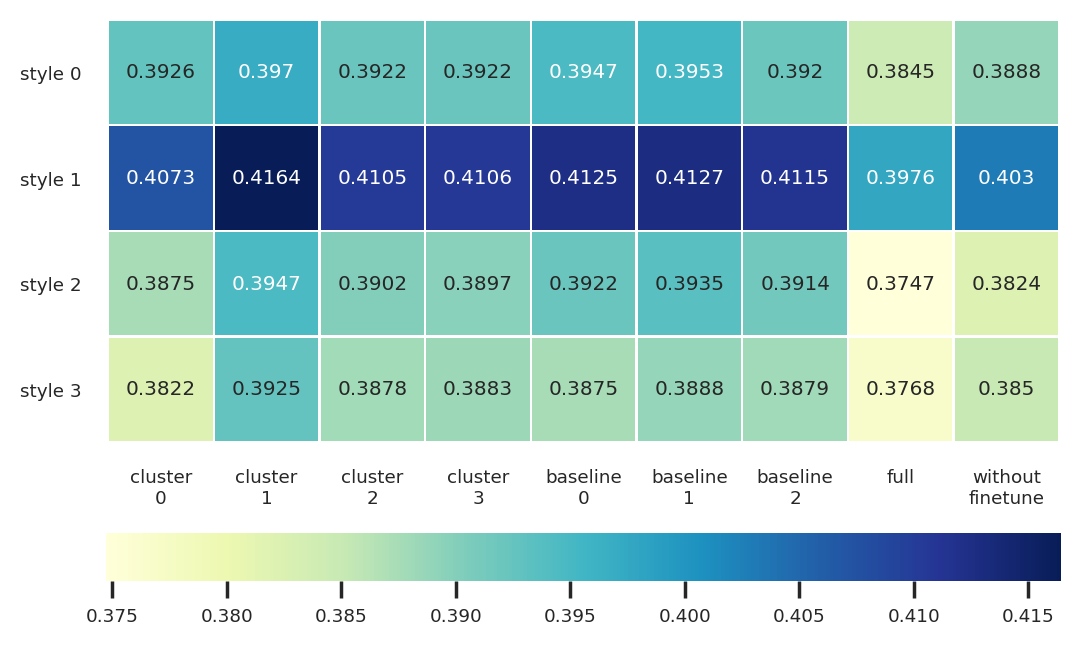}
  \caption{T5-small result(ROUGE-1) on divided test set.}
  \label{fig:test_result}
  \end{minipage}%
\end{figure}

Since the reduction in training data amount brings a drop in model performance, the pretraining model is tested on this task. T5 is a sequence to sequence pretrained language model and can be transferred to the summarization task. A small version of released T5 model is finetuned on both clustered and baseline datasets. What's more, the results that finetuning on the full training set (full) and without finetuning are reported too. The updating steps of finetuning on the full training set are four times the clustered or baselines to ensure that in every set of experiments each sample contributes the same times to the gradient descending.
\begin{table}[htbp]
  \centering
  \small
  \begin{tabular}{|c|c|c|c|c|c|c|c|c|c|c|}
  \hline
  \multirow{2}{*}{} &
    \multicolumn{6}{c|}{\textbf{Summary Related}} &
    \multicolumn{3}{c|}{\textbf{Article Related}} &
    \textbf{NLU} \\ \cline{2-11} 
   &
    \textbf{R1} &
    \textbf{R2} &
    \textbf{RL} &
    \textbf{MTR} &
    \textbf{GLEU} &
    \textbf{BERT} &
    \textbf{N1} &
    \textbf{N2} &
    \textbf{JS} &
    \textbf{Oracle Hit} \\ \hline
  \textbf{Baseline\_0}       & 39.9 & 17.2 & 26.1 & 37.2 & 15.8 & 86.6 & 4.9 & 17.4 & 56.9 & 34.6 \\ \hline
  \textbf{Baseline\_1}       & 40   & 17.2 & 26.2 & 37.2 & 15.8 & 86.8 & 4.8 & 17.6 & 57.4 & 34.8 \\ \hline
  \textbf{Baseline\_2}       & 39.8 & 17   & 26   & 37.1 & 15.7 & 86.7 & 4.9 & 19.6 & 56.2 & 34.5 \\ \hline
  \textbf{Cluster\_0}        & 39.5 & 16.9 & 25.7 & 36.9 & 15.4 & 86.7 & 4.7 & 19.3 & 58.7 & 33.8 \\ \hline
  \textbf{Cluster\_1}        & 40.2 & 17.3 & 26.5 & 37.0   & 16.1 & 86.7 & 4.8 & 18.4 & 57.3 & 35.1 \\ \hline
  \textbf{Cluster\_2}        & 39.7 & 16.9 & 25.7 & 37.0   & 15.6 & 86.8 & 5.1 & 17.9 & 54.0   & 34.3 \\ \hline
  \textbf{Cluster\_3}        & 39.7 & 17.0   & 25.9 & 36.9 & 15.6 & 86.7 & 5.0   & 18.6 & 55.9 & 34.5 \\ \hline
  \textbf{Cluster\_best}     & 43.0   & 20.0   & 30.2 & 40.4 & 18.0   & 87.3 & 4.8 & 17.8 & 57.1 & 38.5 \\ \hline
  \textbf{Without\_finetune} & 39.1 & 16.1 & 25.0   & 32.4 & 15.2 & 86.7 & 4.1 & 8.8  & 79.7 & 36.4 \\ \hline
  \textbf{Full dataset}      & 38.5 & 16.3 & 24.5 & 36.1 & 13.7 & 86.1 & 7.5 & 19.8 & 47.3 & 32.9 \\ \hline
  \end{tabular}
  \caption{Results on CNN-DM dataset.}\label{table:cnndm_result}
  \end{table}

  As shown in Figure \ref{fig:t5_result}, cluster 0 performs best during the learning process, and finetuning on the full dataset is relatively hard to converge. Figure \ref{fig:t5_result} only reports the first 11000 steps of full finetuning. But even it is finetuned until it runs four times as the others it still performs worst. Table \ref{table:cnndm_result} gives much information: results among clusters are consistent with those on PGNet and Transformer. Among various models' results there exist certain styles which are easier to learn and have better generalization (on the full test set). Three baselines perform nearly the same as cluster 1. Hence a single style is not always better than mixed styles for training models. Baseline 1 performs the best because it contains the most samples from cluster 1. Baseline 2 performs the worst since samples don't have clear styles (furthest from cluster centroids). Certain single style is better than mixed clear styles, and mixed clear styles is better than mixed unclear styles and other "bad" single style.

  Another interesting result is that models without finetuning perform better than the finetuned model (full dataset). There may be two reasons: first, the model is pretrained on a multitask target so it has the summarization ability on CNN-DM dataset; second, due to the Language Modeling target the model tends to copy article sentences instead of generating new sentences, which is proved by its poor score on Article Related metrics. So the Summary Related metrics of the original model are higher than those of the finetuned model but actually, the generated summaries are less abstractive. At last, results of "Cluster\_Best" are reported which combines all four clusters and chooses the best summary for each sample. It can be seen as an ensemble model. The result achieves 20.0 on ROUGE-2 F value which is close to the performance (20.1) of the four base-T5 (220M parameters) ensemble finetuned on the full dataset. Although it is not a practicable training trick since there are no gold results in the real test scenario, it suggests that mixing all styles is not the best practice for building an abstractive summarization dataset.

  Figure \ref{fig:test_result} shows the experiment of style's influence on the test set. Each test sample is classified to a cluster by finding the nearest cluster centroid so models' test results on samples with different styles are examined. Surprisingly, the model trained and tested on the same style does not always perform best. The second row and second column in the heatmap show that training data with style 1 leads to the best model and test data with style 1 also scores the best among all models. This may suggest that the current sequence to sequence models prefer the certain settings of summarization tasks instead of all possible patterns from humans.

\section{Conclusion}
In this paper we propose an easy and efficient method to extract summarization subjective styles and give a detailed report on the styles and how it would affect the deep summarization models. To be done......



\bibliographystyle{coling}
\bibliography{coling2020}
\end{document}